\documentstyle[11pt, colacl]{article}

\title{Compiling Language Models from a Linguistically Motivated Unification Grammar}

\author{Manny Rayner$^{1}$, Beth Ann Hockey$^{1}$, Frankie James$^{1}$\\
{\bf Elizabeth Owen Bratt}$^{2}$, {\bf Sharon Goldwater}$^{2}$ {\bf and Jean Mark Gawron}$^{2}$\\
\\
(1) Research Institute for Advanced Computer Science\\
Mail Stop 19-39, NASA Ames Research Center, Moffett Field, CA 94035-1000\\
\\
(2) SRI International, 333 Ravenswood Ave, Menlo Park, CA 94025}

\begin{document}

\maketitle

\begin{abstract}

Systems now exist which are able to compile unification grammars into
language models that can be included in a speech recognizer, but it is so
far unclear whether non-trivial linguistically principled grammars can
be used for this purpose. We describe a series of experiments which
investigate the question empirically, by incrementally constructing a
grammar and discovering what problems emerge when successively larger
versions are compiled into finite state graph representations and used
as language models for a medium-vocabulary recognition task.

\end{abstract}

\section{Introduction}

Construction of speech recognizers for medium-vocabulary dialogue
tasks has now become an important practical problem. The central task
is usually building a suitable language model, and a number of
standard methodologies have become established.  Broadly speaking,
these fall into two main classes. One approach is to obtain or create a
domain corpus, and from it induce a statistical language model,
usually some kind of N-gram grammar; the alternative is to manually
design a grammar which specifies the utterances the recognizer will
accept. There are many theoretical reasons to prefer the first course
if it is feasible, but in practice there is often no choice. Unless a
substantial domain corpus is available, the only method that stands a
chance of working is hand-construction of an explicit grammar based on
the grammar-writer's intuitions.

If the application is simple enough, experience shows that good
grammars of this kind can be constructed quickly and efficiently using
commercially available products like ViaVoice SDK
or the Nuance Toolkit \cite{Nuance}. Systems of this kind typically
allow specification of some restricted subset of the class of
context-free grammars, together with annotations that permit the
grammar-writer to associate semantic values with lexical entries and
rules. This kind of framework is fully adequate for small grammars. As
the grammars increase in size, however, the limited expressive power
of context-free language notation becomes increasingly burdensome. The
grammar tends to become large and unwieldy, with many rules appearing
in multiple versions that constantly need to be kept in step with each
other. It represents a large development cost, is hard to maintain,
and does not usually port well to new applications.

It is tempting to consider the option of moving towards a more
expressive grammar formalism, like unification grammar, writing the
original grammar in unification grammar form and compiling it down to
the context-free notation required by the underlying toolkit. At least
one such system (Gemini; \cite{MooreEtAl97}) has been implemented and
used to build successful and non-trivial applications, most notably
CommandTalk \cite{StentEtAl99}. Gemini accepts a slightly
constrained version of the unification grammar formalism originally
used in the Core Language Engine \cite{Alshawi92}, and compiles it
into context-free grammars in the GSL formalism supported by the
Nuance Toolkit. The Nuance Toolkit compiles GSL grammars
into sets of probabilistic finite state graphs (PFSGs), which form
the final language model.

The relative success of the Gemini system suggests a new
question. Unification grammars have been used many times to build
substantial general grammars for English and other natural languages,
but the language model oriented grammars so far developed for Gemini
(including the one for CommandTalk) have all been domain-specific. One
naturally wonders how feasible it is to take yet another step in the
direction of increased generality; roughly, what we want to do is
start with a completely general, linguistically motivated grammar,
combine it with a domain-specific lexicon, and compile the result down
to a domain-specific context-free grammar that can be used as a
language model. If this programme can be realized, it is easy to
believe that the result would be an extremely useful methodology for
rapid construction of language models. It is important to note that
there are no obvious theoretical obstacles in our way. The claim that
English is context-free has been respectable since at least the early
80s \cite{PullumGazdar82}\footnote{We are aware that this claim is
most probably not true for natural languages in general
\cite{BresnanEtAl87}, but further discussion of this point is beyond
the scope of the paper.}, and the idea of using unification grammar as
a compact way of representing an underlying context-free language is
one of the main motivations for GPSG \cite{GazdarEtAl85} and other
formalisms based on it. The real question is whether the goal is {\it
practically} achievable, given the resource limitations of current
technology.

In this paper, we describe work aimed at the target outlined above, in
which we used the Gemini system (described in more detail in
Section~\ref{Section:Gemini}) to attempt to compile a variety of
linguistically principled unification grammars into language
models. Our first experiments (Section~\ref{Section:ATIS-grammar})
were performed on a large pre-existing unification grammar. These were
unsuccessful, for reasons that were not entirely obvious; in order to
investigate the problem more systematically, we then conducted a
second series of experiments
(Section~\ref{Section:incremental-grammar}), in which we incrementally
built up a smaller grammar. By monitoring the behavior of the
compilation process and the resulting language model as the grammar's
coverage was expanded, we were able to identify the point at which
serious problems began to emerge
(Section~\ref{Section:incremental-grammar-experiments}). In the final
section, we summarize and suggest further directions.

\section{The Gemini Language Model Compiler}
\label{Section:Gemini}

To make the paper more self-contained, this section provides some
background on the method used by Gemini to compile unification
grammars into CFGs, and then into language models. The basic idea is
the obvious one: enumerate all possible instantiations of the features
in the grammar rules and lexicon entries, and thus transform each rule
and entry in the original unification grammar into a set of rules in
the derived CFG. For this to be possible, the relevant features must
be constrained so that they can only take values in a finite
predefined range. The finite range restriction is inconvenient for
features used to build semantic representations, and the formalism
consequently distinguishes {\em syntactic} and {\em semantic}
features; semantic features are discarded at the start of the
compilation process.

A naive implementation of the basic method would be impractical for
any but the smallest and simplest grammars, and considerable ingenuity
has been expended on various optimizations. Most importantly,
categories are expanded in a demand-driven fashion, with information
being percolated both bottom-up (from the lexicon) and top-down (from
the grammar's start symbol). This is done in such a way that
potentially valid combinations of feature instantiations in rules are
successively filtered out if they are not licensed by the top-down and
bottom-up constraints. Ranges of feature values are also kept together
when possible, so that sets of context-free rules produced by the
naive algorithm may in these cases be merged into single rules.

By exploiting the structure of the grammar and lexicon, the
demand-driven expansion method can often effect substantial reductions
in the size of the derived CFG. (For the type of grammar we consider
in this paper, the reduction is typically by a factor of over 10$^{20}$. The downside is that even an apparently small change in the
syntactic features associated with a rule may have a large effect on
the size of the CFG, if it opens up or blocks an important percolation
path. Adding or deleting lexicon entries can also have a significant
effect on the size of the CFG, especially when there are only a small
number of entries in a given grammatical category; as usual, entries
of this type behave from a software engineering standpoint like
grammar rules.

The language model compiler also performs a number of other
non-trivial transformations. The most important of these is related to
the fact that Nuance GSL grammars are not allowed to contain
left-recursive rules, and left-recursive unification-grammar rules
must consequently be converted into a non-left-recursive form.
Rules of this type do not however occur in the grammars described
below, and we consequently omit further description of the
method.

\section{Initial Experiments}
\label{Section:ATIS-grammar}

Our initial experiments were performed on a recent unification grammar
in the ATIS (Air Travel Information System) domain, developed as a
linguistically principled grammar with a domain-specific lexicon.
This grammar was created for an experiment comparing coverage and
recognition performance of a hand-written grammar with that of
automatically derived recognition language models, as increasing
amounts of data from the ATIS corpus were made available for each
method.  Examples of sentences covered by this grammar are ``yes'',
``on friday'', ``i want to fly from boston to denver on united
airlines on friday september twenty third'', ``is the cheapest one
way fare from boston to denver a morning flight'', and ``what flight
leaves earliest from boston to san francisco with the longest layover
in denver''. Problems obtaining a working recognition grammar from the
unification grammar ended our original experiment prematurely, and led
us to investigate the factors responsible for the poor recognition
performance.

We explored several likely causes of recognition trouble: number of
rules, number of vocabulary items, size of node array, perplexity, and
complexity of the grammar, measured by average and highest number of
transitions per graph in the PFSG form of the grammar.  

We were able to immediately rule out simple size metrics as the cause
of Nuance's difficulties with recognition.  Our smallest air travel
grammar had 141 Gemini rules and 1043 words, producing a Nuance
grammar with 368 rules.  This compares to the CommandTalk grammar,
which had 1231 Gemini rules and 1771 words, producing a Nuance grammar
with 4096 rules.

To determine whether the number of the words in the grammar or the
structure of the phrases was responsible for the recognition problems,
we created extreme cases of a Word+ grammar (i.e.\ a grammar that
constrains the input to be any sequence of the words in the
vocabulary) and a one-word-per-category grammar. We found that both of
these variants of our grammar produced reasonable recognition, though
the Word+ grammar was very inaccurate.  However, a
three-words-per-category grammar could not produce successful speech
recognition.

Many feature specifications can make a grammar more accurate, but will
also result in a larger recognition grammar due to multiplication of
feature values to derive the categories of the context-free
grammar. We experimented with various techniques of selecting features
to be retained in the recognition grammar. As described in the
previous section, Gemini's default method is to select only syntactic
features and not consider semantic features in the recognition
grammar.  We experimented with selecting a subset of syntactic
features to apply and with applying only semantic sortal features, and
no syntactic features.  None of these grammars produced successful
speech recognition.

From these experiments, we were unable to isolate any simple set of
factors to explain which grammars would be problematic for speech
recognition. However, the numbers of transitions per graph in a PFSG
did seem suggestive of a factor. The ATIS grammar had a high of 1184
transitions per graph, while the semantic grammar of CommandTalk had a
high of 428 transitions per graph, and produced very reasonable speech
recognition.

Still, at the end of these attempts, it became clear that we did not
yet know the precise characteristic that makes a linguistically
motivated grammar intractable for speech recognition, nor the best way
to retain the advantages of the hand-written grammar approach while
providing reasonable speech recognition.

\section{Incremental Grammar Development}
\label{Section:incremental-grammar}

In our second series of experiments, we incrementally developed a new
grammar from scratch. The new grammar is basically a scaled-down and
adapted version of the Core Language Engine grammar for English
\cite{Pulman92,Rayner93}; concrete development work and testing were
organized around a speech interface to a set of functionalities
offered by a simple simulation of the Space Shuttle
\cite{RaynerEtAl2000}. Rules and lexical entries were added in small
groups, typically 2--3 rules or 5--10 lexical entries in one
increment. After each round of expansion, we tested to make sure that
the grammar could still be compiled into a usable recognizer, and at
several points this suggested changes in our implementation strategy.
The rest of this section describes the new grammar in more detail.

\subsection{Overview of Rules}

The current versions of the grammar and lexicon contain 58 rules and
301 uninflected entries respectively. They cover the following
phenomena:

\begin{enumerate}

\item Top-level utterances: declarative clauses, WH-questions, Y-N
questions, imperatives, elliptical NPs and PPs, interjections.

\item WH-movement of NPs and PPs.

\item The following verb types: intransitive, simple transitive, PP
complement, modal/auxiliary, -ing VP complement, particle+NP
complement, sentential complement, embedded question complement.

\item PPs: simple PP, PP with postposition (``ago''), PP modification
of VP and NP.

\item Relative clauses with both relative NP pronoun (``the temperature
that I measured'') and relative PP (``the deck where I am''). 

\item Numeric determiners, time expressions, and postmodification of NP
by numeric expressions.

\item Constituent conjunction of NPs and clauses.

\end{enumerate}

The following example sentences illustrate current coverage:
{ ``yes''},  
{ ``how about scenario three?''}, 
{ ``what is the temperature?''}, 
{ ``measure the pressure at flight deck''}, 
{ ``go to the crew hatch and close it''}, 
{ ``what were temperature and pressure at fifteen oh five?''}, 
{ ``is the temperature going up?''}, 
{ ``do the fixed sensors say that the pressure is decreasing?''}, 
{ ``find out when the pressure reached fifteen p s i''}, 
{ ``what is the pressure that you measured?''}, 
{ ``what is the temperature where you are?''}, 
{ ``can you find out when the fixed sensors say the temperature at flight deck reached thirty degrees celsius?''}.

\subsection{Unusual Features of the Grammar}

Most of the grammar, as already stated, is closely based on the Core
Language Engine grammar. We briefly summarize the main divergences
between the two grammars.

\subsubsection{Inversion}

The new grammar uses a novel treatment of inversion, which is partly
designed to simplify the process of compiling a feature grammar into
context-free form. The CLE grammar's treatment of inversion uses a
movement account, in which the fronted verb is moved to its notional
place in the VP through a feature. So, for example, the sentence ``is
pressure low?'' will in the original CLE grammar have the
phrase-structure

\vskip 8pt
\noindent   ``[[is]$_{V}$ [pressure]$_{NP}$ [[]$_{V}$ [low]$_{ADJ}$]$_{VP}$]$_{S}$''
\vskip 8pt

\noindent in which the head of the VP is a V gap coindexed with the fronted
main verb ``is''.

Our new grammar, in contrast, handles inversion without movement, by
making the combination of inverted verb and subject into a VBAR
constituent. A binary feature {\tt invsubj} picks out these VBARs, and
there is a question-formation rule of the form

\begin{quote}
{\tt S $\rightarrow$ VP:[invsubj=y]}
\end{quote}

Continuing the example, the new grammar assigns this sentence the
simpler phrase-structure

\vskip 8pt
 \noindent  ``[[[is]$_{V}$ [pressure]$_{NP}$]$_{VBAR}$ [[low]$_{ADJ}$]$_{VP}$]$_{S}$''
\vskip 8pt

\subsubsection{Sortal Constraints}

Sortal constraints are coded into most grammar rules as syntactic
features in a straight-forward manner, so they are available to the
compilation process which constructs the context-free grammar, and
ultimately the language model. The current lexicon allows 11 possible
sortal values for nouns, and 5 for PPs.

We have taken the rather non-standard step of organizing the rules for
PP modification so that a VP or NP cannot be modified by two PPs of
the same sortal type. The principal motivation is to tighten the
language model with regard to prepositions, which tend to be
phonetically reduced and often hard to distinguish from other function
words. For example, without this extra constraint we discovered that an
utterance like

\begin{quote}
   measure temperature at flight deck and lower deck
\end{quote}

\noindent would frequently be misrecognized as

\begin{quote}
   measure temperature at flight deck in lower deck
\end{quote}

\section{Experiments with Incremental Grammars}
\label{Section:incremental-grammar-experiments}

Our intention when developing the new grammar was to find out just
when problems began to emerge with respect to compilation of language
models. Our initial hypothesis was that these would probably become
serious if the rules for clausal structure were reasonably elaborate;
we expected that the large number of possible ways of combining modal
and auxiliary verbs, question formation, movement, and sentential
complements would rapidly combine to produce an intractably loose
language model. Interestingly, this did not prove to be the
case. Instead, the rules which appear to be the primary cause of
difficulties are those relating to relative clauses. We describe the
main results in Section~\ref{Section:relative-clauses}; quantitative
results on recognizer performance are presented together in
Section~\ref{Section:quantitative-results}.

\subsection{Main Findings}
\label{Section:relative-clauses}

We discovered that addition of the single rule which allowed relative
clause modification of an NP had a drastic effect on recognizer
performance. The most obvious symptoms were that recognition became
much slower and the size of the recognition process much larger,
sometimes causing it to exceed resource bounds. The false reject rate
(the proportion of utterances which fell below the recognizer's
minimum confidence theshold) also increased substantially, though we
were surprised to discover no significant increase in the word error
rate for sentences which did produce a recognition result. To
investigate the cause of these effects, we examined the results of
performing compilation to GSL and PFSG level. The compilation
processes are such that symbols retain mnemonic names, so that it is
relatively easy to find GSL rules and graphs used to recognize phrases
of specified grammatical categories.

At the GSL level, addition of the relative clause rule to the original
unification grammar only increased the number of derived Nuance rules
by about 15\%, from 4317 to 4959. The average size of the rules
however increased much more\footnote{GSL rules are written in an
notation which allows disjunction and Kleene star.}. It is easiest to
measure size at the level of PFSGs, by counting nodes
and transitions; we found that the total size of all the graphs had
increased from 48836 nodes and 57195 transitions to 113166 nodes and
140640 transitions, rather more than doubling. The increase was not
distributed evenly between graphs. We extracted figures for only the
graphs relating to specific grammatical categories; this showed that the
number of graphs for NPs had increased from 94 to 258, and moreover that
the average size of each NP graph had increased from 21 nodes and 25.5
transitions to 127 nodes and 165 transitions, a more than sixfold
increase. The graphs for clause (S) phrases had only increased in number
from 53 to 68. They had however also greatly increased in average
size, from 171 nodes and 212 transitions to 445 nodes and 572
transitions, or slightly less than a threefold increase. Since NP and
S are by far the most important categories in the grammar, it is not
strange that these large changes make a great difference to the quality
of the language model, and indirectly to that of speech recognition.

Comparing the original unification grammar and the compiled GSL
version, we were able to make a precise diagnosis. The problem with
the relative clause rules are that they unify feature values in the
critical S and NP subgrammars; this means that each constrains the
other, leading to the large observed increase in the size and
complexity of the derived Nuance grammar. Specifically, agreement
information and sortal category are shared between the two daughter
NPs in the relative clause modification rule, which is schematically
as follows:
\begin{quote}
{\tt NP:[agr=A, sort=S] $\rightarrow$\\
 NP:[agr=A, sort=S]\\
 REL:[agr=A, sort=S]}
\end{quote}
These feature settings are needed in order to get the right alternation
in pairs like
\begin{quote}
the robot that *measure/measures the temperature [agr]\\
\\
the *deck/temperature that you measured [sort]
\end{quote}
We tested our hypothesis by commenting out the {\tt agr} and {\tt
sort} features in the above rule. This completely solves the main
problem of explosion in the size of the PFSG representation; the new
version is only very slightly larger than the one with no relative
clause rule (50647 nodes and 59322 transitions against 48836 nodes and
57195 transitions). Most importantly, there is no great increase in
the number or average size of the NP and S graphs. NP graphs increase
in number from 94 to 130, and stay constant in average size; S graphs
increase in number from 53 to 64, and actually decrease in average
size to 135 nodes and 167 transitions. Tests on speech data show that
recognition quality is nearly the same as for the version of the
recognizer which does not cover relative clauses. Although 
speed is still significantly degraded, the process size has been
reduced sufficiently that the problems with resource bounds disappear.

It would be reasonable to expect that removing the explosion in the
PFSG representation would result in an underconstrained language model
for the relative clause part of the grammar, causing degraded
performance on utterances containing a relative clause. Interestingly,
this does not appear to happen, though recognition speed under the new
grammar is significantly worse for these utterances compared to
utterances with no relative clause.

\subsection{Recognition Results}
\label{Section:quantitative-results}

This section summarizes our empirical recognition results.  With the
help of the Nuance Toolkit {\tt batchrec} tool, we evaluated three
versions of the recognizer, which differed only with respect to the
language model. {\tt no\_rels} used the version of the language model
derived from a grammar with the relative clause rule removed; {\tt
rels} is the version derived from the full grammar; and {\tt unlinked}
is the compromise version, which keeps the relative clause rule but
removes the critical features. We constructed a corpus of 41
utterances, of mean length 12.1 words. The utterances were chosen so
that the first 31 were within the coverage of all three versions of
the grammar; the last 10 contained relative clauses, and were within
the coverage of {\tt rels} and {\tt unlinked} but not of {\tt
no\_rels}. Each utterance was recorded by eight different subjects,
none of whom had participated in development of the grammar or
recognizers. Tests were run on a dual-processor SUN Ultra60 with 1.5 GB
of RAM.

The recognizer was set to reject utterances if their associated
confidence measure fell under the default
threshold. Figures~\ref{Table:no-rel-results}
and~\ref{Table:rel-results} summarize the results for the first 31
utterances (no relative clauses) and the last 10 utterances (relative
clauses) respectively. Under `xRT', we give mean recognition speed
(averaged over subjects) expressed as a multiple of real time; `FRej'
gives the false reject rate, the mean percentage of utterances which
were rejected due to low confidence measures; `Mem' gives the mean
percentage of utterances which failed due to the recognition process
exceeding memory resource bounds; and `WER' gives the mean word error
rate on the sentences that were neither rejected nor failed due to
resource bound problems. Since the distribution was highly skewed, all
means were calculated over the six subjects remaining after exclusion
of the extreme high and low values.

Looking first at Figure~\ref{Table:no-rel-results}, we see that {\tt
rels} is clearly inferior to {\tt no\_rels} on the subset of the
corpus which is within the coverage of both versions: nearly twice as
many utterances are rejected due to low confidence values or resource
problems, and recognition speed is about five times slower. {\tt
unlinked} is in contrast not significantly worse than {\tt no\_rels} in
terms of recognition performance, though it is still two and a half
times slower. 

Figure~\ref{Table:rel-results} compares {\tt rels} and {\tt unlinked}
on the utterances containing a relative clause. It seems reasonable
to say that recognition performance is comparable for the two versions:
{\tt rels} has lower word error rate, but also rejects more utterances.
Recognition speed is marginally lower for {\tt unlinked}, though it is
not clear to us whether the difference is significant given the high
variability of the data.

\begin{figure}
\begin{center}
\begin{tabular}{|c|r|r|r|r|}
\hline
Grammar          &  xRT &  FRej   &  Mem   & WER   \\
\hline
\hline
{\tt no\_rels}   & 1.04 &  9.0\%  &  --    & 6.0\% \\  
\hline
{\tt rels}       & 4.76 &  16.1\% &  1.1\% & 5.7\% \\ 
\hline
{\tt unlinked}   & 2.60 &   9.6\% &  --    & 6.5\% \\
\hline
\end{tabular}
\caption{Evaluation results for 31 utterances not containing relative
clauses, averaged across 8 subjects excluding extreme values.}
\label{Table:no-rel-results}
\end{center}
\end{figure}

\begin{figure}
\begin{center}
\begin{tabular}{|c|r|r|r|r|}
\hline
Grammar          &  xRT  &  FRej   &  Mem   &  WER   \\
\hline
\hline
{\tt rels}       &  4.60 &  26.7\%  &  1.6\% &  3.5\% \\ 
\hline
{\tt unlinked}   &  5.29 &  20.0\%  &  --    &  5.4\% \\
\hline
\end{tabular}
\caption{Evaluation results for 10 utterances containing relative
clauses, averaged across 8 subjects  excluding extreme values.}
\label{Table:rel-results}
\end{center}
\end{figure}

\section{Conclusions and Further Directions}

We found the results presented above surprising and interesting. When
we began our programme of attempting to compile increasingly larger
linguistically based unification grammars into language models, we had
expected to see a steady combinatorial increase, which we guessed
would be most obviously related to complex clause structure. This did
not turn out to be the case. Instead, the serious problems we
encountered were caused by a small number of critical rules, of which
the one for relative clause modification was by the far the
worst. It was not immediately obvious how to deal with the
problem, but a careful analysis revealed a reasonable compromise solution,
whose only drawback was a significant but undisastrous degradation
in recognition speed.

It seems optimistic to hope that the relative clause problem is the
end of the story; the obvious way to investigate is by continuing to
expand the grammar in the same incremental fashion, and find out what
happens next. We intend to do this over the next few months, and
expect in due course to be able to present further results.

\section{Acknowledgements}

The research described in Section 3 was supported by the Defense
Advanced Research Projects Agency under Contract N66001--94--C--6046
with the Naval Command, Control, and Ocean Surveillance Center. The majority of the research reported was performed at RIACS under NASA Cooperative Agreement Number NCC 2-1006.

\end{document}